%% file: main.tex
\documentclass[10pt,conference]{IEEEtran}
\usepackage[T1]{fontenc} 
\usepackage[hidelinks]{hyperref}
\usepackage{amsmath}
\usepackage{nicefrac}
\usepackage{tikz}
\usetikzlibrary{tikzmark}
\usepackage{booktabs}
\usepackage{mathtools} 
\usepackage{amsfonts}
\usepackage{longtable}
\usepackage{booktabs}
\usepackage{bm}
\usepackage{siunitx}
\usepackage{algorithm}
\usepackage{standalone}
\usepackage{enumerate}
\usepackage[capitalize]{cleveref}
\crefmultiformat{equation}{Eqs.~(#2#1#3)}%
{ or~(#2#1#3)}{, (#2#1#3)}{ or~(#2#1#3)}
\usepackage[noEnd=true]{algpseudocodex}

\usepackage{vector}
\renewcommand{\vec}[1]{\vect{#1}}

\algrenewcommand\algorithmicfunction{\textbf{Function}}
\usepackage[style=ieee, doi=false,isbn=false,url=false,eprint=false,backend=biber]{biblatex}

\AtEveryBibitem{\clearfield{month}}

\DeclareFieldFormat{year}{\thefield{year}}
\AtEveryBibitem{%
  \iffieldundef{year}
    {}
    {\clearfield{month}%
     \clearfield{day}%
     \clearfield{endmonth}%
     \clearfield{endday}}%
}

\AtEveryCitekey{%
  \iffieldundef{year}
    {}
    {\clearfield{date}}%
}

\renewbibmacro*{date}{%
  \ifboolexpr{
    test {\iffieldundef{year}}
  }
  {}
  {\printtext{\printfield{year}}}
}
\AtEveryBibitem{\clearfield{series}}

\DeclareBibliographyDriver{unpublished}{%
  \usebibmacro{bibindex}%
  \usebibmacro{begentry}%
  \usebibmacro{author/translator+others}%
  \newunit\newblock
  \usebibmacro{title}%
  \newunit\newblock
  \printfield{note}%
  \newunit\newblock
  \printfield{eprinttype}%
  \newunit
  \printfield{eprint}%
  \newunit\newblock
  \usebibmacro{date}%
  \newunit\newblock
  \usebibmacro{finentry}%
}

\DeclareBibliographyDriver{online}{%
  \usebibmacro{bibindex}%
  \usebibmacro{begentry}%
  \usebibmacro{author/translator+others}%
  \newunit\newblock
  \usebibmacro{title}%
  \newunit\newblock
  \printfield{note}%
  \newunit\newblock
  \printfield{eprinttype}%
  \newunit
  \printfield{eprint}%
  \newunit\newblock
  \usebibmacro{date}%
  \newunit\newblock
  \usebibmacro{finentry}%
}
\AtEveryBibitem{\clearfield{pagetotal}}

\input{preamble/pgfplots_preamble}

\newcommand{\bcalN}{{\mathcal{N}}}
\newcommand{\bxi}{\vect{\xi}}
\newcommand{\bmu}{\vect{\mu}}
\newcommand{\bbeta}{\vect{\beta}}
\newcommand{\bbetah}{\bar{\bbeta}}
\newcommand{\setI}{\mathcal{I}}

\newcommand{\bx}{\vect{x}}
\newcommand{\by}{\vect{y}}
\newcommand{\br}{\vect{r}}
\newcommand{\bw}{\vect{w}}
\newcommand{\bv}{\vect{v}}
\newcommand{\beps}{\vect{\varepsilon}}
\newcommand{\xiyiset}{
{\{(\bx_i, \by_i)\}}_{i \in \setI}
}
\newcommand{\xiyisetRV}{
{\{(\bar{\bx}_i, \bar{\by}_i)\}}_{i \in \setI}
}
\newcommand{\betaposterior}{p_{\bbetah \mid \xiyisetRV}}
\newcommand{\xih}{\bar{\xi}}
\newcommand{\bxih}{\bar{\bxi}}

\newcommand{\Iset}{\mathcal{I}}
\newcommand{\overbar}[1]{\mkern 4.0mu\overline{\mkern-4.0mu#1\mkern-4.0mu}\mkern 4.0mu} 

\title{Probabilistic Parameter Estimators and Calibration Metrics for Pose Estimation from Image Features}

\author{\IEEEauthorblockN{Romeo Valentin\IEEEauthorrefmark{1}\IEEEauthorrefmark{4}, Sydney M. Katz\IEEEauthorrefmark{1}, Joonghyun  Lee\IEEEauthorrefmark{1}, \\ Don Walker\IEEEauthorrefmark{2}, Matthew Sorgenfrei\IEEEauthorrefmark{2}, and Mykel J. Kochenderfer\IEEEauthorrefmark{1}}
\IEEEauthorblockA{\IEEEauthorrefmark{1}Stanford Intelligent Systems Laboratory, Stanford University, Stanford, CA, 94305\\
\IEEEauthorrefmark{2}A$^3$ by Airbus LLC, Sunnyvale, CA, 94086\\
\IEEEauthorrefmark{4}Email: {romeov@stanford.edu}\\
}}

\begin{document}

\maketitle

\begin{abstract}
	This paper addresses the challenge of probabilistic parameter estimation given measurement uncertainty in real-time.
	We provide a general formulation and apply this to pose estimation for an autonomous visual landing system.
	We present three probabilistic parameter estimators: a least-squares sampling approach, a linear approximation method, and a probabilistic programming estimator.
	To evaluate these estimators, we introduce novel closed-form expressions for measuring calibration and sharpness specifically for multivariate normal distributions.
	Our experimental study compares the three estimators under various noise conditions.
	We demonstrate that the linear approximation estimator can produce sharp and well-calibrated pose predictions significantly faster than the other methods but may yield overconfident predictions in certain scenarios.
	Additionally, we demonstrate that these estimators can be integrated with a Kalman filter for continuous pose estimation during a runway approach where we observe a 50\% improvement in sharpness while maintaining marginal calibration.
	This work contributes to the integration of data-driven computer vision models into complex safety-critical aircraft systems and provides a foundation for developing rigorous certification guidelines for such systems.
\end{abstract}

\begin{IEEEkeywords}
	Parameter Estimation, Uncertainty Quantification, Calibration,
	Pose Estimation,
	Computer Vision, Probabilistic Programming

\end{IEEEkeywords}

\input{sections/introduction}
\input{sections/preliminaries}
\input{sections/three_estimators}
\input{sections/measuring_calibration}

\input{sections/experiments}
\input{sections/conclusion}
\input{sections/acknowledgements}
\printbibliography
\input{sections/appendix}

\end{document}

%% file: preamble/pgfplots_preamble.tex
\usepackage{tikz}
\usepackage{pgfplots}

\usepgfplotslibrary{statistics}
\usepackage{nicefrac}
\pgfplotsset{compat=1.18}
\usetikzlibrary{positioning}
\usepackage[T1]{fontenc}
\usepackage{xcolor}
\usepackage{siunitx}
\usepackage{bm}

\colorlet{bgcolor}{black!4}
\colorlet{darkgreen}{green!50!black}

%% file: sections/introduction.tex
\section{Introduction}\label{introduction}

Automation in aviation has recently gained technological advancements based on data-driven models applied to
vision, decision-making, planning, and human
interaction.
A direct ``proof of correctness'' or ``learning
assurance'' for such models currently seems far away, making practitioners
and regulators resort to other approaches \cite{EASA2024AI,harveyHowFDARegulates2020a,balduzziNeuralNetworkBased,corsoHolisticAssessmentReliability2023,valentinFrameworkDeepLearning2024}.
These approaches may include test set validation~\cite{bishop2006pattern}, out-of-distribution detection~\cite{yangGeneralizedOutofDistributionDetection2024}, causal models~\cite{pearlCausality2009,petersElementsCausalInference2017,scholkopfCausalityMachineLearning2022}, mechanistic interpretability~\cite{liEmergentWorldRepresentations2022,nandaProgressMeasuresGrokking2023,templeton2024scaling}, adversarial robustness~\cite{goodfellowExplainingHarnessingAdversarial2015}, uncertainty estimation~\cite{kendallWhatUncertaintiesWe2017,liuSimplePrincipledUncertainty2020,gawlikowskiSurveyUncertaintyDeep2023}, and calibration~\cite{gneitingProbabilisticForecastsCalibration2007,guoCalibrationModernNeural2017,angelopoulosGentleIntroductionConformal2022a}.

In this work, we focus on the principled processing and validation of model output uncertainty.
First, we consider a general setting for probabilistic parameter estimation from uncertain measurements.
Then, we consider the pose estimation problem arising as part of an autonomous visual landing system.
We assume that the system relies on a data-driven computer vision algorithm that finds the exact projection location of
runway corners in the image, and has known reference coordinates.
Further, we assume the model dynamically produces a quantification of uncertainty in its predictions.

\newpage
This paper makes the following contributions:
\begin{enumerate}[I.]
	\item
	      Three formulations for general uncertainty-aware parameter estimation,
	      and their specific formulation for the pose estimation problem;
	\item
	      A novel closed-form expression for measuring calibration and sharpness
	      for multivariate normal distributions; and
	\item
	      An experimental study comparing the three estimators for pose
	      estimation using calibration and sharpness metrics.
\end{enumerate}

The paper is organized as follows. \cref{sec:preliminaries} provides some necessary background and notation.
\cref{sec:solver-formulations} first proposes three estimators for probabilistic parameter estimation based on
(i) repeated noise sampling and least-squares minimization, (ii) a linear approximation,
and (iii) a Bayesian approach solved using Markov Chain Monte Carlo, and then discusses integrating the proposed estimators into a Kalman filter.

\cref{subsec:measuring-calibration} introduces techniques to evaluate the
quality of multivariate probabilistic predictions through (i) calibration and (ii) sharpness.
In particular, we propose closed-form expressions for efficiently computing calibration
and sharpness for multivariate normal distributions given ground truth observations, which allow us to
compare the efficacy of the proposed estimators.

\cref{sec:experiments} presents experimental results for all three estimators applied to an aircraft pose estimation problem during a runway approach under different noise conditions.
We find that the linear approximation estimator typically constructs sharp and calibrated predictions two orders of magnitude faster than the other estimators.
However, we also find that it produces overconfident predictions in particular scenarios.
We further demonstrate integration of the estimators with a Kalman filter for a single approach, which leads to a $2\times$ improvement in sharpness, while maintaining marginal calibration.

Overall, this work aims to improve understanding of how to best integrate data-driven computer vision models into complex safety-critical aircraft, and further provide a foundation upon which rigorous
certification guidelines for such systems can be built.
Code supporting this work is available online.\footnote{
	See \url{https://github.com/sisl/RunwayPNPSolve.jl}, \\
	\url{https://github.com/RomeoV/ProbabilisticParameterEstimators.jl}, and \\
	\url{https://github.com/RomeoV/MvNormalCalibration.jl}.}

%% file: sections/preliminaries.tex
\section{Preliminaries}\label{sec:preliminaries}

We briefly introduce necessary notation and phrase pose estimation as a general parameter estimation problem.
This section reviews least squares estimation given measurement uncertainty.
We then discuss how to represent measurement and pose uncertainty, and how to evaluate whether
a series of predictions is (i) probabilistically ``correct'' (calibrated) and (ii) precise (sharp).

\begin{figure}[t]
	\centering
	\includegraphics[width=\linewidth]{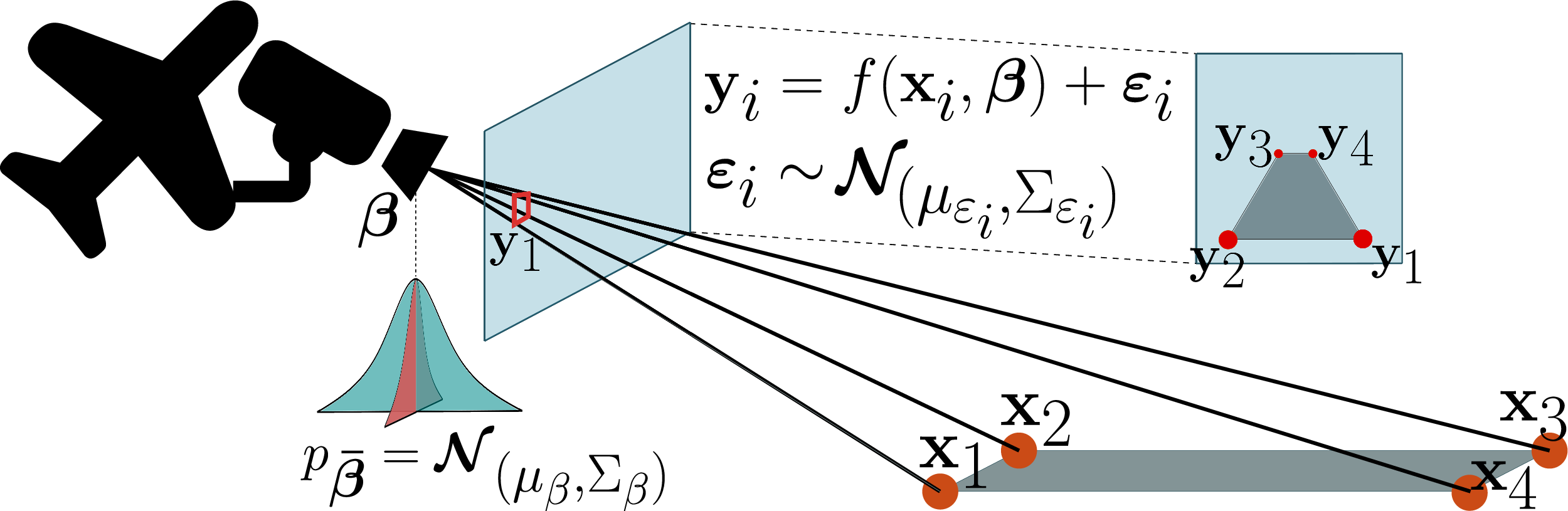}
	\caption{Known world points \(\bx_{i}\) are projected onto a
		camera at \(\bbeta\), where they are measured as
		\(\by_{i}\) under the influence of noise. We wish to determine
		\(\bbeta\) or a distribution over
		\(\bbeta\).
	}\label{fig:overview}
\end{figure}

\subsection{Camera Model for Point Projections}
Consider the pose estimation problem for an aircraft approaching a runway approach using image features with known corresponding world points.
We assume a (pinhole) camera located on the aircraft, positioned such
that the entire runway is in view of the camera lens.
\Cref{fig:overview} presents an overview.

Let the camera projection function \(\by = f(\bx, \bbeta)\)
map a world point
$\bx = \lbrack x_{\lbrack \text{alongtrack}\rbrack}, x_{\lbrack \text{crosstrack} \rbrack}, x_{\lbrack \text{altitude} \rbrack}\rbrack$
to an image projection
$\by = \lbrack y_{\lbrack \text{right} \rbrack}, y_{\lbrack \text{up} \rbrack} \rbrack$
given a camera position
$\bbeta = \lbrack \beta_{\lbrack \text{alongtrack} \rbrack}, \beta_{\lbrack \text{crosstrack} \rbrack}, \beta_{\lbrack \text{altitude} \rbrack}\rbrack$

Consider now a focal-point-centric coordinate system rotated such that
the $x$-axis goes from the focal point through the center of the camera
projection plane towards the world points, and the $y$- and $z$-axes are
parallel to the projection plane's edges.
If we write the location of the world points in this coordinate system as
\(\lbrack x^\prime, y^\prime, z^\prime\rbrack\), we can compute the
camera projection coordinate simply as
\begin{equation}\by = \lambda_{\text{focal length}} \cdot \frac{1}{x^\prime}\begin{bmatrix}
		y^\prime \\
		z^\prime
	\end{bmatrix}.
	\label{eq:projection1}
\end{equation}
We will use the notation
\begin{equation}
	\by_{i} = \begin{bmatrix}
		\left( y_{i} \right)_{\lbrack 1\rbrack} \\
		\left( y_{i} \right)_{\lbrack 2\rbrack}
	\end{bmatrix},\quad i \in \Iset
\end{equation}
to denote the first and second
components of the \(i\)-th observation \(\by_{i}\), with
\(\Iset = \left\{ 1,2,3,4 \right\}\) for the four runway corners.
We denote random variables with a hat and write
\begin{equation}{\xih}^{(k)} \sim p_{\xih},\end{equation} to denote the \(k\)-th sample
of the random variable \(\xih\)
that follows the distribution \(p_{\xih}\).
Upright bold variables such as \(\by\) denote vectors, and matrices are uppercase such as \(\Sigma\).




\subsection{Pose-from-N-Points (PNP)}

To solve the PNP problem in its basic form, it is now sufficient to
solve the nonlinear least-squares problem
\begin{equation}\bbeta = \arg\min\limits_{\bbeta}\sum_{i}{d\left( \by_{i},f\left( \bx_{i},\bbeta \right) \right)}^{2}
	\label{eq:lsq1}
\end{equation}
where \(d( \cdot , \cdot )\) is the Euclidean distance.
For other problem settings, $d$ and $f$ can take a variety of forms.
For example, if $f$ computes sidelines angles other geometric features, such as sideline angles,
an appropriate distance functions may be chosen.
We also point out that for point projections, due to the nonlinearity of the projection function \(f( \cdot , \cdot )\),
this problem is numerically challenging to solve.
In particular, for low altitudes, the inverse problem becomes very sensitive to small changes in
the projection. Nonetheless, for approach angles of at least
\(\qty{1}{\degree}\) above ground, the PNP problem can typically be solved
using, e.g., the Newton-Raphson, Levenberg-Marquardt, or trust region
algorithms.

\subsection{Least-squares Under Measurement Noise}\label{sec:weighted-pnp}

If weights \(w_{i}\) for each observation are available, we may rewrite
\cref{eq:lsq1} as
\begin{equation}\bbeta = \arg\min\limits_{\bbeta}\sum_{i}w_{i} \cdot {d\left( \by_{i},f\left( \bx_{i},\bbeta \right) \right)}^{2}
	\label{eq:lsq2}
\end{equation}
where weights are typically chosen as the inverse variance of each
observation, i.e., \(w_{i} = 1/{\sigma_{y}^{2}}_{i}\) \cite{zhangParameterEstimationTechniques1997}.
For noise that is
correlated across observations, we can instead concatenate all components
into a single vector
\begin{equation}{\br}(\bbeta) = \begin{bmatrix}
		d\left( \by_{1},f\left( \bx_{1},\bbeta \right) \right) \\
		d\left( \by_{2},f\left( \bx_{2},\bbeta \right) \right) \\
		\vdots
	\end{bmatrix}
	\label{eq:residual_vector}
\end{equation} where each
component \(d( \cdot , \cdot )\) may be further expanded if it has
multiple components (such as component-wise distances). Then we can rewrite
\cref{eq:lsq2} as
\begin{equation}\bbeta = \arg\min\limits_{\bbeta}{\br(\bbeta)}^\top W\br(\bbeta)
	\label{eq:lsq3}
\end{equation}
where \(W\) denotes the weight matrix which can be chosen as the inverse
variance-covariance matrix of the observation noise, e.g.
\(\Sigma_{\by}^{- 1}\) for the case of Gaussian noise \cite{zhangParameterEstimationTechniques1997}.

Notice that we can also rewrite \cref{eq:lsq3} as \begin{equation}
	\begin{aligned}
		\bbeta & = \arg\min\limits_{\bbeta}{\tilde{\br}}(\bbeta)^\top \tilde{\br}(\bbeta) \\
		       & = \arg\min\limits_{\bbeta}\sum_{i}{\tilde{r}}_{i}^{2}(\bbeta)
	\end{aligned}
	\label{eq:lsq4}
\end{equation} by LU-factorizing
\(W = LU\) and setting
\(\tilde{\br} = U{\br}\) in
the general case, or \(\Sigma = LU\) and
\(\tilde{\br} =  L^{-\top} {\br}\)
in the Gaussian noise case. This representation is sometimes preferable
for mathematical convenience and solver interfaces.

\subsection{Representing Measurement and Pose Uncertainty}

If our world points
\(\left\{ \bx_{i} \right\}_{i \in \Iset}\) and measurements
\(\left\{ \by_{i} \right\}_{i \in \Iset}\) are known, we
can solve the pose estimation problem as a nonlinear least squares
optimization as shown above. However, our measurements
\(\by_{i}\) and world points \(\bx_{i}\) can be noisy or
uncertain, and may therefore be provided through a probabilistic
description, e.g., a Gaussian or other distribution, or samples drawn
from a distribution.

In this scenario, our pose estimate will be a random variable \(\bbetah\), or more precisely, \(\bbetah \mid \xiyisetRV\). We therefore need to represent the pose either (i) through samples (to which a distribution may be fitted) or (ii)
directly as a (multivariate) distribution. In these cases, we write
\begin{equation}
	\left( {\bbetah}^{(1)},{\bbetah}^{(2)},\ldots \right)\overset{\text{ i.i.d.}}{\sim}\betaposterior
	\label{eq:pr-beta1}
\end{equation}
with realizations
\(\left( \bbeta^{(1)},\bbeta^{(2)},\ldots \right)\), or
\begin{equation}\betaposterior \approx \bcalN_{(\mu,\Sigma)},
	\label{eq:pr-beta2}
\end{equation}
respectively. Notice that due to the non-linearity in the projection in
\cref{eq:projection1}, the distribution
\(\betaposterior\)
will not in general be Gaussian, even for Gaussian noise in the measurements. However, we
will see that approximating
\(\betaposterior\)
as a Gaussian \(\bcalN_{(\mu,\Sigma)}\) does yield good results for
our problem.

\subsection{Calibration and
	Sharpness}\label{subsec:measuring-calibration-prelims}

Calibration gives us a way to quantify whether the uncertainty in a
series of predictions is faithful to the distribution of prediction
errors. Given a set of predicted probability distributions
\(p_{{\xih}_{i}}\) and corresponding measurements \(\xi_{i}\), we
call the predictions \emph{(marginally) calibrated} if
\begin{equation}\text{Pr}\left( \xih_{i} \leq q(p_{\xih_{i}}, \rho) \right) \approx \rho\quad\forall\rho \in (0,1),
	\label{eq:marginal-calibration}
\end{equation}
where \(q(p_{\xih_{i}}, \rho)\) denotes the quantile
function for the random variable \({\xih}_{i}\) evaluated at
\(\rho\). Notably, however, the quantile function is not defined for
multivariate distributions such as
\(\betaposterior\)
in \cref{eq:pr-beta1} and \cref{eq:pr-beta2}. We must therefore
generalize \cref{eq:marginal-calibration} to remove the dependence on
the quantile function and instead rely on the more general notion of
prediction sets. We present such a solution in
\cref{subsec:measuring-calibration}.

Further, we note that a series of predictions may be perfectly
calibrated yet still be ``bad'' in the sense that each prediction has
large uncertainty. Therefore, we must also measure sharpness, a measure
of the ``conciseness'' of the predictions. We propose a measure of
sharpness for multivariate normal distributions in
\cref{subsec:measuring-sharpness} and refer to
\textcite{gneitingProbabilisticForecasting2014} for an overview of the
trade off between calibration and sharpness. 




%% file: sections/three_estimators.tex
\section{Three Probabilistic Estimators}\label{sec:solver-formulations}

Given an observation function \(f( \cdot , \cdot )\) and measurement
tuples \(\xiyiset\), we have seen how to estimate the parameter \(\bbeta\),
incorporating measurement uncertainties through a weighting scheme given in
\cref{sec:weighted-pnp}. However, this process will only give us the maximum likelihood estimator for
\(\bbeta\), i.e.
\begin{equation}
	\bbeta = \arg\max\limits_{\bbeta} \betaposterior(\bbeta).
\end{equation}
Instead, we are now interested in constructing a probabilistic
description of the parameter \(\bbeta\) as a distribution or a
set of samples. We propose three estimators for this task: (i) the
least-squares sampling estimator, (ii) the linear approximation
estimator, and (iii) the MCMC estimator. We show that although the three methods differ in mathematical rigor
and runtime cost, their performance often matches closely.

\subsection{The LSQ Estimator via Noise
	Sampling}\label{subsec:estimator-lsq}

The simplest approach to constructing a sampled representation of
\(p_{\bbetah}\) is to repeatedly sample simulated noise and
solve the least-squares formulation for each noise sample.
To construct the $k$th pose estimate \(\bbeta^{(k)}\),
we draw a noise sample \(\beps_{i}^{(k)}\) for each
corner projection \(\by_{i}\) and solve the least squares
formulations from \cref{eq:lsq4} where we substitute
\begin{equation}
	\by_{i} \leftarrow \by_{i} - \beps_{i}^{(k)}
\end{equation}
for each \(i\). Repeating this a number of times will result in a set of estimates
\begin{equation}
	\{ \bbeta^{(k)} \}_{k \in \mathcal{K}} =
	\{ \bbeta^{(1)},\bbeta^{(2)},\ldots,\bbeta^{\left( |\mathcal{K}| \right)} \}
\end{equation}
which we may optionally approximate as a distribution \(\betaposterior\) by fitting a
multivariate normal distribution to the samples.
For the pose estimation application we find that a number of samples \(|\mathcal{K}|\)
between 100 and 500 is typically sufficient.

Notably, the LSQ Estimator does not require us to make any assumptions
about the prior distribution \(p_{\bbetah}\) or the shape of the posterior \(\betaposterior\)
(which is in general not a perfect Gaussian), nor does it require strong assumptions
about the function \(f( \cdot , \cdot )\). Indeed, even multimodal
posterior distributions can be approximated with this method.

The downside of this method is that the computational cost is rather high, as the full
least-squares problem has to be solved for every sample of
\(\bbeta^{(k)}\). However, starting from the second sample,
setting the initial guess for \(\bbeta^{(k)}\) to the solution
of \(\bbeta^{(k - 1)}\) can significantly reduce the runtime cost.

\subsection{Linear Approximation Estimator}

The LSQ Estimator suffers from high computational cost, since it has to
solve the least squares problem many times. However, if we approximate
the projection function \(f\left( \bx,\bbeta \right)\) as
a linear function around some sample \(\bbeta^{(1)}\), and
assume the noise to be Gaussian, we can construct
\(p_{\bbetah}\) in a single step by directly propagating
the uncertainty in the observations \(\by\).
We call this estimator the Linear Approximation Estimator.

Let \(J_{ij} = \frac{\partial f_{i}\left( \bx,\bbeta \right)}{\partial\beta_{j}}\)
be the Jacobian of our projection function. We can relate changes in
\(\bbeta\) and \(\br\) or \(\widetilde{\br}\) as
\begin{equation}\begin{aligned}
		J\Delta\bbeta                                    & = - \Delta\widetilde{\br}, \\
		\Rightarrow \left( J^\top WJ \right)\Delta\bbeta & = - J^\top W\Delta\br.
	\end{aligned}
	\label{eq:deltabeta_deltar}
\end{equation}
With
\(\br_{i} = \by_{i} - f\left( \bx,\bbeta \right)\)
and
\(\by_{i} = f\left( \bx_{i},\bbeta \right) + \beps_{i}\)
we can model \(\Delta\br = - \beps\) and therefore
treat \(\Delta\bbeta\) as a random variable
\(\overbar{\Delta\bbeta}\) with distribution
\begin{equation}
	p_{\overbar{\Delta\bbeta}} = \left( J^\top WJ \right)^{- 1}J^\top W{p_{\bar{\beps}}}_{i}.
\end{equation}
Using this insight, we can now construct our parameter estimate
\(p_{\bbetah}\) as follows.
We first estimate \(\mu_{\bbeta}\) by solving \cref{eq:lsq4} once,
without any noise perturbation.
Then, we compute
\begin{equation}\begin{aligned}
		p_{\bbetah} & = \mu_{\bbeta} + p_{\overbar{\Delta\bbeta}}                            \\
		            & = \mu_{\bbeta} + \left( J^\top WJ \right)^{- 1}J^\top Wp_{\bar{\beps}}
	\end{aligned}
	\label{eq:linearmodel1}
\end{equation}
which yields the desired result.
For example, if
\(p_{\bar{\beps}} = \bcalN_{\left( \vec{0},\Sigma_{\beps} \right)}\)
and \(W = \Sigma_{\beps}^{- 1}\) then
\begin{equation}
	p_{\bbetah} = \bcalN_{\left( \mu_{\bbeta},\Sigma_{\bbeta} \right)}
\end{equation}
with
\begin{gather}
	\mu_{\bbeta} = \arg\min\limits_{\bbeta}{\br(\bbeta)}^\top \br(\bbeta) \\
	\Sigma_{\bbeta} = \left( J^\top \Sigma_{\beps}^{- 1}J \right)^{- 1}J^\top \Sigma_{\beps}^{- 1}J\left( J^\top \Sigma_{\beps}^{- 1}J \right)^{- 1}.
\end{gather}
Further, if \(J\) is square and invertible then
\(\Sigma_{\bbeta}\) reduces to
\begin{equation}
	\Sigma_{\bbeta} = J^{- 1}\Sigma_{\beps}\left( J^\top  \right)^{- 1}.
\end{equation}
Finally, we may replace \(p_{\bar{\beps}}\) in \cref{eq:linearmodel1} with
\(\left( p_{\bar{\by}} - f\left( \bx,\theta^{\text{est}} \right) \right)\)
to slightly relax the assumption that
\(\by_{i} = f\left( \bx_{i},\bbeta \right) + \beps_{i}\),
which can yield more robust results.

The Linear Approximation Estimator
is highly efficient, only requiring to solve the least squares problem
once and compute the Jacobian of \(f\) one additional time. This property makes
it more suitable for real-time systems such as a visual positioning
system. However, in cases where the assumption of Gaussian noise does
not hold (see e.g. \cref{subsec:results-heavy-tail-noise}), or if the
function is highly nonlinear, the estimator may produce poor results.

\subsection{MCMC Estimator}

The previous two approaches have relied on a least-squares formulation
of the problem, carefully constructing a weighting scheme to incorporate
uncertainties correctly. Now, we present an approach based on a
Bayesian formulation that can create samples \(\bbeta^{(k)}\)
directly using probabilistic programming.

Our formulation is as follows. We choose a weakly informative prior
\(p_{\bbetah}\) as a diagonal multivariate normal
distribution such that our entire operational domain is contained within
one standard deviation of the prior. Of course, if a better prior is
available it may be chosen instead. We further recall that we assume we
have a model of the observation noise \({p_{\beps}}_{i}\)
(which is now not constrained to any particular family of distribution),
provided by our sensor for every sample.
Then our Bayesian formulation is simply
\begin{equation}\begin{aligned}
		\bbetah         & \sim p_{\bbetah}                                            \\
		{\bar{\by}}_{i} & \sim f\left( \bx_{i},\bbeta \right) + {p_{\bar{\beps}_{i}}}
	\end{aligned}
	\label{eq:mcmc2}
\end{equation} which can be directly
solved using Markov Chain Monte Carlo \cite{metropolisEquationStateCalculations1953,gelmanBayesianDataAnalysis1995}, yielding samples
\(\left( \bbeta^{(1)},\bbeta^{(2)},\ldots \right)\).

This formulation is remarkably simple, necessitating no further thoughts
about how to integrate the knowledge of uncertainties into any kind of
weighting scheme, how to set up the least squares problem, or
assumptions about the distributions of inputs or outputs. Further, this
formulation can indeed be more computationally efficient than the noise
sampling approach from \cref{subsec:estimator-lsq} (see e.g.
\cref{tab:runtime-characteristics}) and can represent arbitrary distribution shapes just like the LSQ Estimator.


\subsection{Integrating Probabilistic Estimators Into a Kalman Filter Framework}\label{subsec:kalman-filter-integration}
Finally, we demonstrate how we can integrate the proposed probabilistic parameter estimators into a Kalman filter framework.
Recall the Kalman filter equations (without inputs) given by
\begin{equation}\begin{aligned}
		\bx^{[t + 1]} & = A\bx^{[t]}     + \bw^{[t]}     \\
		\by^{[t + 1]} & = C\bx^{[t + 1]} + \bv^{[t + 1]}
	\end{aligned}\end{equation}
with process noise
\(\mathbb{E}\left\lbrack \bw^{[ t^{\prime} ]}{\bw^{[t]}}^\top  \right\rbrack = \delta_{t^{\prime}t}Q^{[t]}\)
and measurement noise
\(\mathbb{E}\left\lbrack \bv^{[ t^{\prime} ]}{\bv^{[t]}}^\top  \right\rbrack = \delta_{t^{\prime}t}R^{[t]}\),
i.e., Gaussian noise uncorrelated in time with covariances \(Q^{[t]}\) and \(R^{[t]}\) \cite{kalman1960new}.

Often it is difficult to choose $R^{[t]}$ dynamically for each time step, and instead a single fixed $R$ is chosen
for all time steps.
Using one of the proposed estimators, however, we can dynamically compute $\by^{[t]}$ and $R^{[t]}$ for each time step.
%
Consider the probabilistic parameter $\bbetah^{[t]}$ to represent our measurements $\by^{[t]}$ at time step $t$, with
\begin{equation}
	p_{\bbetah^{[t]} \mid {\{(\bar{\bx}^{[t]}_i, \bar{\by}^{[t]}_i)\}}_{i \in \setI}} \approx \bcalN_{(\mu^{[t]}, \Sigma^{[t]})}.
\end{equation}
Then,
for each time step, we can set
\begin{equation}
	\begin{aligned}
		\by^{[t]} & \leftarrow \bmu^{[t]}    \\
		R^{[t]}   & \leftarrow \Sigma^{[t]}.
	\end{aligned}
\end{equation}

Notice how this formulation differs from typical nonlinear Kalman filter formulations such as the Extended Kalman filter \cite{kalman1960new,kalmanNewResultsLinear1961}.
Our measurement uncertainty is not restricted to relations $\by = h(\bx) + \bw$ for some known $h(\bx)$, but instead may arise without modeling $h(\bx)$, such as measured image features with predictive uncertainty perceived from state $\bx$.

One important additional consideration is that the Kalman filter assumes
noise uncorrelated between time steps, which is typically not the case for high-frequency
sensors, such as a computer vision model.
We consider two possible solutions:
(i)
Use only measurements far enough apart in time that their error correlation can be assumed to be low; or
(ii) use a Kalman filter formulation that is able to deal with ``colored''
noise.
We refer to \textcite{wangPracticalApproachesKalman2012} for an overview of the latter.

%% file: sections/measuring_calibration.tex
\begin{algorithm}[t]
	\caption{point\_in\_pset(\ldots)}
	\begin{algorithmic}[1]
		\Statex \textbf{Inputs:} Distribution $\bcalN_{(\mu,\Sigma)}$,
		\Statex \phantom{\textbf{Inputs:}} Sample $\bxi$,
		\Statex \phantom{\textbf{Inputs:}} Target coverage rate $\rho$;
		\Statex \textbf{Output:} Is $\bxi \text{ in } \mathcal{P}_{\rho}(\bcalN_{(\mu, \Sigma)})$?
		\Function{point\_in\_pset}{$\bcalN_{(\mu,\Sigma)}, \bxi, \rho$}
		\State $Q \gets \text{eigvecs}(\Sigma)$
		\State $\tilde{\bxi} \gets Q^T(\bxi - \bmu)$
		\State $\tilde{\Sigma} = \text{diag}\left([\tilde{\sigma}^2_{[1]}, ..., \tilde{\sigma}^2_{[d]}]\right) \gets Q^T\Sigma Q$
		\State $d \gets \text{dimension}(\bxi)$
		\State \Return all(
		\State \hspace{1em} \textsc{point\_in\_pset\_1d}($\mathcal{N}_{(0,\tilde{\sigma}^2_{[l]})}, \tilde{\xi}_{[l]}, \rho$)
		\State \hspace{1em} for $l \in \{1, ..., d\}$)
		\EndFunction
		\State
		\Function{point\_in\_pset\_1d}{$\mathcal{N}_{(\mu,\sigma^2)}, \xi, \rho$}
		\State \Return $|\frac{\xi-\mu}{\sigma}| < \text{quantile}\left(\mathcal{N}_{(0,1)}, \frac{\rho}{2} + \frac{1}{2}\right)$
		\EndFunction
	\end{algorithmic}
	\label{alg:point-in-pset}
\end{algorithm}

\section{Measuring Calibration and Sharpness for Multivariate Normal Distributions}\label{subsec:measuring-calibration}

Evaluating the quality of a probabilistic estimator is not trivial.
For example, consider predictions somewhat close to the truth but overly confident and predictions
further from the truth but with appropriately high uncertainty.
The latter is often preferable due to it not misleading us with high confidence.
Conversely, predictions that are ``probabilistically correct'' but include an overly large number of possibilities can be equally problematic.

These two properties have been formalized as ``calibration'' and ``sharpness'' and are crucial for assessing and comparing the reliability of estimators.
In \cref{sec:experiments} we will see that, depending on the problem setup, some of the proposed estimators can be sharp but not calibrated, or calibrated but not sharp, or even marginally calibrated in each component but not jointly calibrated.
However, as introduced in \cref{subsec:measuring-calibration-prelims}, computing calibration is typically restricted to univariate distributions due to its reliance on the quantile or cumulative density function, and therefore not applicable to our predictions \(p_{\bbetah}\).

To remedy this, in this section we introduce efficient methods for measuring calibration and sharpness specifically tailored for multivariate normal distributions through the use of ``centered prediction sets'' in a diagonalizing basis.

\subsection{Centered Prediction Sets}

In this section, we construct a scheme to efficiently compute
calibration for multivariate normal distributions using centered
prediction sets \(\mathcal{P}_{\rho}\).
Consider for now the univariate probability distribution \(p_{\xih}\) of a random variable \(\xih\) and a
sample \(\xih_i\) drawn from \(p_{\xih}\).
Then we define a prediction set
\(\mathcal{P}_{\rho}( p_{\xih} )\) at coverage level
\(\rho\) such that the probability of \(\xih_i\) being contained in
\(\mathcal{P}_{\rho}( p_{\xih} )\) is approximately
equal to \(\rho\), i.e.
\begin{equation}\text{ Pr}\left( \xih_i \in \mathcal{P}_{\rho}( p_{\xih} ) \right) \approx \rho.
	\label{eq:prediction-set}
\end{equation}
We can see we recover \cref{eq:marginal-calibration} exactly
by setting
\begin{equation}\label{eq:exact-recovery-pset}
	\mathcal{P}_\rho(p_{\xih}) = \bigl( - \infty,\ q(p_{\xih},\rho)\bigr\rbrack.
\end{equation}
However, we can also choose different constructions.
For example, we can also choose the centered construction
\begin{equation}\label{eq:centered-pset}
	\mathcal{P}_\rho = \left[q(p_{\xih}, \nicefrac{1}{2} - \nicefrac{\rho}{2}),\ q(p_{\xih}, \nicefrac{1}{2} + \nicefrac{\rho}{2})\right]
\end{equation}
or even the disjoint (and centered) construction
\begin{align}\label{eq:disjoint-pset}
	\mathcal{P}_{\rho}( p_{\xih} ) = \phantom{\cup}\  & \bigl( - \infty,q(p_{\xih},\nicefrac{\rho}{2})\bigr\rbrack \\ \cup\ &\bigl\lbrack q(p_{\xih},1 - \nicefrac{\rho}{2}),\infty\bigr).
\end{align}
Notice that each construction satisfies \cref{eq:prediction-set} but typically differs significantly in size.
Specifically, if $p_{\xih}$ is Gaussian then \cref{eq:centered-pset} constructs the smallest set to satisfy \cref{eq:prediction-set} for any $\rho$ because it accumulates always those points with highest probability density.
We will call this construction \emph{centered} and \emph{cumulative},
as opposed to \emph{off-centered} such as \cref{eq:exact-recovery-pset} or \emph{disjoint} such as \cref{eq:disjoint-pset}.

In this work we will proceed by considering only the centered and cumulative construction from \cref{eq:centered-pset}.
Although the other constructions would also suffice for our upcoming multivariate construction,
we find that the central and cumulative construction more accurately reflects a small calibration error given a small mean prediction error and is further symmetric about the direction of the error.


\begin{algorithm}[t]
	\caption{compute\_calibration(\dots)}
	\begin{algorithmic}[1]
		\Statex \textbf{Inputs:} Set of predictions $\{\bcalN_i\}_{i\in\Iset}$,
		\Statex \phantom{\textbf{Inputs:}} Set of observations $\{{\bxi}_i\}_{i\in\Iset}$,
		\Statex \phantom{\textbf{Inputs:}} Target coverage rate $\rho$;
		\Statex \textbf{Output:} Empirical coverage rate.
		\Function{compute\_calib}{$\{\bcalN_i\}_{i\in\Iset}, \{\bxi_i\}_{i\in\Iset}, \rho$}
		\State mean(
		\State \hspace{1em} \textsc{point\_in\_pset}($\bcalN, \bxi, \rho$)
		\State \hspace{1em} for $(\bcalN, \bxi) \in \text{zip}(\{\bcalN_i\}_{i\in\Iset}, \{\bxi_i\}_{i\in\Iset})$
		\State )
		\EndFunction
	\end{algorithmic}
	\label{alg:compute-calibration}
\end{algorithm}

\subsection{Constructing Cumulative Centered Prediction Sets for
	Multivariate Normal Distributions}

The construction of central and cumulative prediction sets introduced above is still not applicable for multivariate distributions.
However, by making two key observations about multivariate normal distributions we will be able to efficiently construct
prediction sets $\mathcal{P}(\bcalN_{(\mu, \Sigma)})$ that satisfy \cref{eq:prediction-set} just as above.

First, any multivariate normal distribution
\(\bcalN_{(\mu,\Sigma)}\) with $d$ dimensions can be ``diagonalized'' without
fundamentally changing the probability density by considering a new distribution
\(\bcalN_{\left( \widetilde{\mu},\widetilde{\Sigma} \right)}\)
in a rotated basis such that the covariance matrix
\begin{align}
	\widetilde{\Sigma} & = Q\Sigma Q^\top                                                                                                                                       \\
	                   & = \text{ diag}\left( \left\lbrack {\widetilde{\sigma}}_{\lbrack 1\rbrack}^{2},\ldots,{\widetilde{\sigma}}_{\lbrack d\rbrack}^{2} \right\rbrack \right)
\end{align}
is diagonalized with a rotation matrix \(Q\) and
appropriately transformed samples
\(\widetilde{\bxi} = Q^\top \left( \bxi - \bmu \right)\)
(and thus
\(\widetilde{\bmu} = Q^\top \left( \bmu - \bmu \right) = \bm{0}\)).

Second, the prediction set
\(\mathcal{P}_{p}(\bcalN_{( \tilde{\mu},\tilde{\Sigma} )})\)
for diagonalized normal distributions can simply be constructed by
considering each dimension individually as a univariate case, with
\begin{equation}
	\begin{aligned}
		\mathcal{P}_{\rho}\left( \bcalN_{( \tilde{\mu},\tilde{\Sigma} )} \right) =
		\bigl\{\widetilde{\bxi}\, \mid \,{\widetilde{\xi}}_{\lbrack i\rbrack} \in \mathcal{P}_{\sqrt[d]{\rho}}( \mathcal{N}_{( 0,{\widetilde{\sigma}}_{\lbrack i\rbrack}^{2} )} ) &         \\
		\forall\left\{ i \in 1,\ldots,d \right\}                                                                                                                                  & \bigr\}
		.
	\end{aligned}
\end{equation}
Notice that we have replaced the probability \(\rho\) of each set with \(\sqrt[d]{\rho}\)
because the probability of $d$ conditions each with probability $\sqrt[d]{\rho}$ to hold simultaneously is exactly $\rho$.

For the univariate case we now recall the construction from \cref{eq:centered-pset} and further exploit the symmetric structure of Gaussians to rewrite \cref{eq:centered-pset} as
\begin{equation}\begin{aligned}
		\mathcal{P}_{\rho}\left( \mathcal{N}_{\left( \mu,\sigma^{2} \right)} \right)
		 & = \left\lbrack q(\mathcal{N}_{\left( \mu,\sigma^{2} \right)},\nicefrac{1}{2} - \nicefrac{\rho}{2}),\ q(\mathcal{N}_{\left( \mu,\sigma^{2} \right)},\nicefrac{1}{2} + \nicefrac{\rho}{2}) \right\rbrack \\
		 & = \left\{ \xi \mathrel{\Big|} \frac{\xi - \mu}{\sigma} < q\left(\mathcal{N}_{\left( 0,1^{2} \right)},\frac{1}{2} + \frac{\rho}{2}\right) \right\}.
	\end{aligned}\end{equation}
Notably, when considering samples normalized by the
prediction's mean and standard deviation, we find that the prediction set can
be written only as a function of \(\rho\), which makes this construction
very efficient. To give an example of the construction, consider
\(p_{\xih} = \mathcal{N}_{(\mu,\Sigma)} = \mathcal{N}_{(\tilde\mu,\tilde\Sigma)}\) with
\(\mu = \lbrack \frac{1}{3}. \frac{1}{4} \rbrack \) and
\(\Sigma = \text{diag}( \lbrack 1.2^{2},1.5^{2}\rbrack )\),
and pick \(\rho = 0.68^{2} \approx 0.46\). Then we can construct
\begin{equation}
	\mathcal{P}_{0.46}( p_{\xih} ) = \left\{ \xi \mathrel{\Big|} \frac{\mspace{9mu}\left( \xi_{\lbrack 1\rbrack} - \frac{1}{3} \right)}{1.2} \leq 1.0 \land \frac{\xi_{\lbrack 2\rbrack} - \frac{1}{4}}{1.5} \leq 1.0 \right\}.
\end{equation}
\cref{alg:point-in-pset} provides an implementation of this construction.


\subsection{Evaluating Calibration}

Given the above constructions of prediction sets for multivariate normal distributions, it
is now straightforward to compute the calibration, or coverage rate, for any \(\rho\), which we will be able to plot as a calibration curve.
This calibration curve will essentially tell us how much \cref{eq:prediction-set} is violated given predictions and actual observations.

Given a sequence of Gaussian predictions
\(\{p_{{\bxih}_{i}}\}_{i \in \Iset}\) and
observations \(\{ {\bxi}_{i} \}_{i \in \Iset}\)
with
\({p_{\bxih}}_{i} = \bcalN_{\left( \mu_{i},\Sigma_{i} \right)}\)
we can compute the ratio of observations contained in the
prediction set \(\mathcal{P}_{\rho}(p_{\bxih})\), i.e.,
\begin{equation}\text{ coverage } \coloneq \frac{1}{|\setI|}\sum_{i \in \Iset}\left\lbrack {\bxih}_{i} \in \mathcal{P}_{\rho}( {p_{\bxih}}_{i} ) \right\rbrack,\end{equation}
where \(\lbrack \ \cdot\  \rbrack\) denotes the Iverson bracket.
\cref{alg:compute-calibration} implements this evaluation scheme.
Plots of this coverage rate are shown in \cref{sec:experiments}.

\subsection{Defining and Computing
	Sharpness}\label{subsec:measuring-sharpness}

Finally, we discuss how to efficiently compute sharpness for
multivariate normal distributions. Unfortunately, sharpness does not
have a single definition comparable to \cref{eq:prediction-set} for
calibration. However, it is still useful for comparing
predictions.

We define sharpness of a single prediction as the
{hyper-volume} of the set of points within one standard deviation of the
mean prediction. Specifically, if we consider again the diagonalized
covariance matrix
\(\widetilde{\Sigma} = \text{ diag}( \lbrack {\widetilde{\sigma}}_{\lbrack 1\rbrack}^{2},\ldots,{\widetilde{\sigma}}_{\lbrack d\rbrack}^{2} \rbrack )\)
we can compute sharpness as
\begin{equation}\label{eq:sharpness} \text{ sharpness } \coloneq V_{d} = \frac{\pi^{d/2}}{\Gamma(d/2 + 1)}\prod_{i}{\widetilde{\sigma}}_{i}\end{equation}
where \(\Gamma( \cdot )\) is the Gamma function. Notice that, for
example, with \(d = 1\) \cref{eq:sharpness} reduces to \(2\widetilde{\sigma}\), and
with \(d = 2\) and
\(\Sigma = \text{ diag}( \lbrack {\widetilde{\sigma}}_{1}^{2},{\widetilde{\sigma}}_{2}^{2} \rbrack )\)
reduces to \(\pi{\widetilde{\sigma}}_{1}{\widetilde{\sigma}}_{2}\).

%% file: sections/experiments.tex
\section{Experiments}\label{sec:experiments}
Finally, we present experimental results computing calibration and sharpness for
camera pose estimates obtained using the three proposed estimators under
different noise conditions.
To this end, we consider randomly sampled
aircraft poses several kilometers away from a typically sized commercial
runway and observe corner projections under noise. These measurements
\(\by_{i}\) and measurement noise estimates \(\Sigma_{\by_{i}}\)
are assumed to be outputs of a sensor, typically a neural network with
uncertainty outputs, see e.g. \textcite{lakshminarayananSimpleScalablePredictive2017}.
We also present an integration study considering a single continuous approach.
We demonstrate how we can filter our pose estimates using a Kalman filter
formulation with the measurement noise in each step given by our
estimators. Details on the experimental setup and estimator parameters are provided in \cref{subsec:runway-details,subsec:lsq-details,subsec:mcmc-details}.

\subsection{Multivariate Uncorrelated Normal Noise}\label{subsec:results-multivariate-uncorrelated-normal-noise}
\begin{figure}[t]
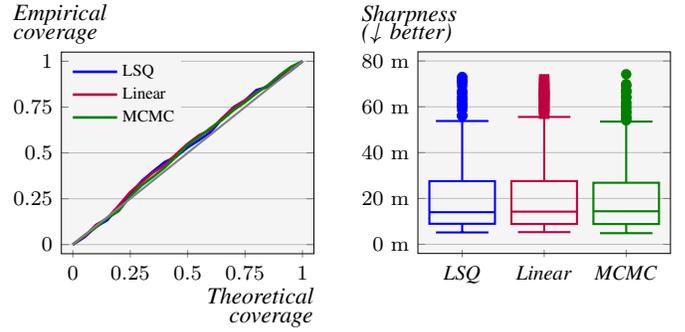

	\centering
	\includestandalone[width=\linewidth]{pgfplots/calib_uncorrelated}
	\vspace{-3.0ex}
	\caption{Uncorrelated normal noise: Calibration and sharpness results.
	}\label{fig:results-uncorrelated-gaussian}
\end{figure}

As our first experiment, we consider multivariate uncorrelated normal noise for each observation with
\begin{equation} \label{eq:uncorrelated-noise-model}
	\begin{gathered}
		\beps_{i} \sim \bcalN_{( \vec{0},\Sigma_{\beps} )}, \\
		\mathbb{E}\left\lbrack \beps_{i}\beps_{j}^\top  \right\rbrack = \vec{0}_{2 \times 2}\quad\forall i,j \in \left\{ 1,\ldots,4 \right\}
	\end{gathered}
\end{equation}
This model  allows for correlation between the ``up'' and ``right'' components of a single corner's projection, but does not correlations between the projections of different corners.
\cref{fig:results-uncorrelated-gaussian} shows the calibration curve and
sharpness distribution for 300 experiment instantiations, i.e. 300 true
poses sampled according to \cref{eq:random-pose} and their corresponding pose estimates.

All three
estimators produce well-calibrated results and have a very similar
sharpness distribution, which may indicate that all three estimators
produce the ``optimal'' solution given the data. We also notice that there are some significant sharpness outliers. In
context, we hypothesize that this means that for some noise samples there is no pose
that explains the observations well, and any model must make a
prediction with high uncertainty. For a runtime system, we therefore may
choose to employ filtering techniques over time, an example of which we
show in \cref{subsec:filtering-results}.

\subsection{Multivariate Correlated Normal Noise}

Next, we verify that the three different weighting strategies for incorporating correlated noise all produce correct results.
For this, we consider multivariate correlated normal noise
\begin{equation}\label{eq:correlated-noise-model}
	\begin{gathered}
		\beps_{i} \sim \bcalN_{\left( \vec{0},\Sigma_{\beps} \right)}, \\
		\mathbb{E}\left\lbrack \beps_{i}\beps_{j}^\top  \right\rbrack = \begin{bmatrix}
			0.7 & 0   \\
			0   & 0.7
		\end{bmatrix}\quad\forall i \neq j,
	\end{gathered}
\end{equation}
i.e., the error of the ``right'' components of projections are correlated
across corners, and the same for the ``up'' components.
This reflects observations we have found in real sensor measurements,
where sometimes errors in components are correlated across observations \(\by_i\).

We find that the results closely resemble the results from the previous section, although with increased sharpness outliers.
We conclude that all three estimators incorporate the correlation terms correctly into their respective weighting schemes, even though the details of the algorithms differs between estimators.

We also contrast these results with another experiment where we sample noise according to \cref{eq:correlated-noise-model} but do not model the correlation terms between observations, i.e., we model the problem according to \cref{eq:uncorrelated-noise-model}.
We find that for all estimators the results (misleadingly) improve sharpness but produce overconfident results, which manifest as calibration curves with a ``U'' shape (we will see a similar result in the next experiment).
These results show the importance of correctly estimating and modeling the noise including the covariate terms.




\subsection{Long Tail Noise}\label{subsec:results-heavy-tail-noise}
Finally, we examine a case where the three estimators differ.
We consider univariate ``long tail'' noise, modeled as a superposition of a low- and a
high-variance Gaussian, with
\begin{equation}\left( \varepsilon_{i} \right)_{\lbrack l\rbrack} \sim \left\lbrack \frac{3}{4}\mathcal{N}_{\left( 0,1^{2} \right)} + \frac{1}{4}\mathcal{N}_{\left( 0,3^{2} \right)} \right\rbrack\quad\forall i,l,\end{equation}
i.e. there are no correlations between measurements or components, but
each component's noise is either sampled from a ``concise'' or ``long
tail'' distribution.
Notice that this noise violates the assumptions of the Linear Approximation Estimator,
which assumes Gaussian noise.

\begin{figure}[t]
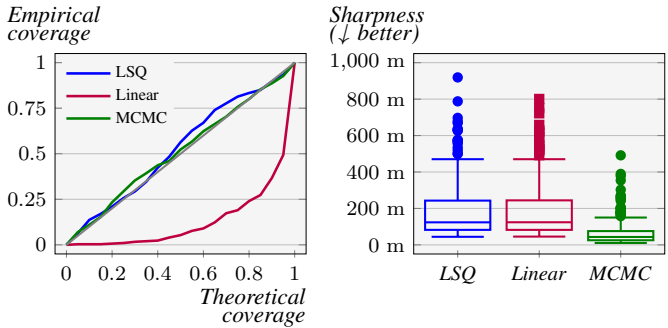

	\centering
	\includestandalone[width=\linewidth]{pgfplots/calib_long_tail}
	\vspace{-3.0ex}
	\caption{Long tail noise: Calibration and sharpness results.
	}\label{fig:results-mixture-model}
\end{figure}

\cref{fig:results-mixture-model} shows the results for this noise model.
Notably, the linear models fails here, producing either overconfident predictions
or mispredicting the mean (we cannot tell from the calibration plot alone).
This indicated that although the linear model can be a
good choice for certain applications due to its computational speed, it
must be used with care with respect to its assumptions.

We further observe that the MCMC and LSQ Estimators are both well-calibrated, but the
prior significantly outperforms the latter in terms of sharpness.
This showcases that an estimator can produce calibrated and sharp results
but may still not be ``ideal'' because another estimator may produce sharper results.

\subsection{Runtime Characteristics}\label{subsec:runtime-characteristics}
One important difference between the estimators is their runtime
characteristics. We recall that the MCMC and LSQ Estimators produce
samples \(\bbeta^{(k)}\) of \(p_{\bbetah}\), and we
must fit a multivariate normal distribution to produce
\(p_{\bbetah}\), which we find requires between 100 and 500
samples of \(\bbetah\) to achieve calibrated and sharp results.
This is in contrast to the Linear Approximation Estimator, which
directly constructs \(p_{\bbetah}\) in a single step.


\begin{table}[t]
	\caption{Runtime characteristics of each estimator.}
	\centering
	\begin{tabular}{@{}l rrrr@{}}
		\toprule
		                        & \tikzmarknode{mymark}{100} & 300    & 500    & $\pm\sigma$ \\
		\cmidrule(lr){2-5}
		\emph{LSQ Est.}         & 54 ms                      & 188 ms & 311 ms & 7\%         \\
		\emph{Lin. Approx Est.} & 0.4 ms                     & 0.4 ms & 0.4 ms & 11\%        \\
		\emph{MCMC Est.}        & 53 ms                      & 105 ms & 183 ms & 6\%         \\
		\bottomrule
	\end{tabular}
	\label{tab:runtime-characteristics}
\end{table}
\tikz[remember picture, overlay] \node[anchor=east, font=\scriptsize, yshift=-0.1ex] at (mymark.west) {$|\mathcal{K}|=$};

In \cref{tab:runtime-characteristics}, we present timing results and
standard measurement error \(\sigma\) given the uncorrelated noise model
from \cref{eq:uncorrelated-noise-model}. Timings were collected using
an 11th Gen Intel Core i7-11800H @ 2.30GHz with 32GB RAM, Julia version
\texttt{1.10.4}, and LLVM version \texttt{15.0.7}. For
any \(k\), the LSQ Estimator takes about \(k\) times as long as the
Linear Approximation Estimator, as can be expected.

We also notice that for \(|\mathcal{K}| = 300\) and \(500\), the MCMC Estimator
manages to perform almost twice as fast as the LSQ Estimator, despite
generating an additional \(250\) warmup samples (see \cref{subsec:mcmc-details}
for details on the warmup).
This can be explained by the different nature of MCMC and LSQ algorithms,
with the MCMC algorithms not having to solve a full minimization problem and instead directly sampling from the posterior.

We conclude that, if the required assumptions hold, the Linear Approximation Estimator outperforms the other two proposed estimators in runtime performance while achieving similar results.
We also suggest that due to its computational speed, the Linear Approximation Estimator in particular may be suitable for integration into a real-time safety-critical system.
When the assumptions of the Linear Approximation Estimator do not hold, we suggest that even the more computationally demanding MCMC Estimator may be considered for such applications.

\subsection{Filtering Predictions Along an
	Approach}\label{subsec:filtering-results}
Finally, we consider a case study of integrating probabilistic pose estimates into a Kalman filter framework.
We consider real sensor data from a single, randomly chosen runway approach to compute probabilistic estimates of our pose at each time step.
We then process the time series with a Kalman filter, using the parameter uncertainty for the Kalman filter's measurement model as introduced in \cref{subsec:kalman-filter-integration}.

For this experiment, we consider a simple uncorrelated Gaussian noise model similar to \cref{subsec:results-multivariate-uncorrelated-normal-noise} and therefore choose the Linear Approximation Estimator.
To assure that the measurement errors are not highly correlated in time, we choose a minimum time step of \(\Delta t = \qty{1}{\second}\), for which we find that the measurement error autocorrelation falls below 1/3 of its maximum.
We choose the state $\bx$ to consist of position and linear velocities. 
The observations $\by$ are then chosen to be only the position, which is estimated by $\bbeta$.

\begin{figure}[t]
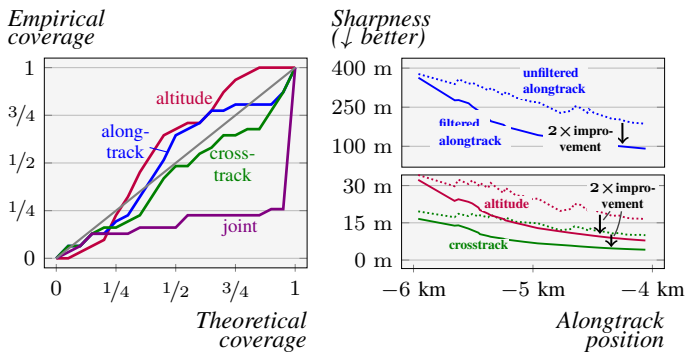

	\centering
	\includestandalone[width=1.043\linewidth]{pgfplots/calib_filtered_x}
	\vspace{-3.0ex}
	\caption{Kalman filter integration: Calibration results for filtered marginal and joint estimates, and sharpness results for both unfiltered (dotted) and filtered (solid) estimates along a single approach.
	}
	\label{fig:filtering-results}
\end{figure}

\cref{fig:filtering-results} presents the results of this experiment.
We can see that filtering brings a \(2 \times\) improvement in sharpness
in all three components (alongtrack, crosstrack, and altitude) while maintaining marginal calibration for each
component individually.
However, interestingly we lose joint calibration, i.e., calibration of the full pose estimate $p_{\bbetah}$, despite the
unfiltered estimates being well-calibrated (not shown here).
We conclude that although the marginal terms are computed correctly, the filtering introduces errors in the covariate terms between the components.
Nonetheless, we conclude that the proposed probabilistic pose estimators can be used with great efficacy for a typical pose estimation problem.


%% file: pgfplots/calib_uncorrelated.tex
\begin{tikzpicture}

	\begin{axis}[
			name=plot1,
			footnotesize,
			enlargelimits=0.05,
			xlabel={\emph{Theoretical}\\[-0.75ex] \emph{coverage}},
			ylabel={\emph{Empirical}\\[-0.75ex] \emph{coverage}},
			x label style={at={(xticklabel cs:1)}, anchor=north east, inner sep=0pt, outer sep=0pt, align=right, scale=1.0},
			y label style={at={(yticklabel cs:1)}, rotate=-90, anchor=south west, align=left, scale=1.0},
			xtick distance=0.25,
			ytick distance=0.25,
			xtick align=center,
			ytick align=inside,
			axis background/.style={fill=bgcolor},
			line width=0pt,
			y tick style={line width=0pt},
			xtick pos=bottom,
			ymajorgrids=true,
			legend style={fill=bgcolor, inner sep=1pt, fill opacity=0.5, text opacity=1, nodes={scale=0.8}, draw=none},
			legend cell align={left},
			legend pos=north west,
		]

		\addplot [ blue, line width=1.0pt,]
		table[x=pvals, y=calibrationvals] {pgfplots/data/uncorrelated_calibration_lsq.dat}
		coordinate[pos=0.5] (foo)
		;
		\addlegendentry{LSQ};

		\addplot [ purple, line width=1.0pt,]
		table[x=pvals, y=calibrationvals] {pgfplots/data/uncorrelated_calibration_lin.dat}
		;
		\addlegendentry{Linear};

		\addplot [ green!50!black, line width=1.0pt,]
		table[x=pvals, y=calibrationvals] {pgfplots/data/uncorrelated_calibration_mcmc.dat}
		;
		\addlegendentry{MCMC};



		\addplot [
			gray,
			opacity=0.5,
			line width=0.8pt,
		]
		coordinates {(0,0) (1,1)};

	\end{axis}

	\begin{axis}[at=(plot1.north east), anchor=left of north west, xshift=1.5em,
			name=plot2,
			y label style={at={(yticklabel cs:1)}, rotate=-90, anchor=south west, align=left, scale=1.0},
			x label style={at={(xticklabel cs:1)}, anchor=north east, inner sep=0pt, outer sep=0pt, align=right, scale=1.0, yshift=-0.33ex},
			ylabel={\emph{Sharpness}\\[-0.75ex] \emph{(\textdownarrow\ better)}},
			enlargelimits=0.05,
			footnotesize,
			boxplot/draw direction=y,
			axis background/.style={fill=bgcolor},
			xtick={1, 2, 3},
			xticklabels={\emph{LSQ}, \emph{Linear}, \emph{MCMC}},
			yticklabel={\pgfmathprintnumber{\tick} $\unit{\m}$},
			x tick label style={rotate=0},
			xtick align=inside,
			ytick align=inside,
			ytick style={line width=0pt},
			ymin=0,
			ymax=80,
			ymajorgrids=true,
			line width=0pt,
			xtick align=center,
			xtick pos=bottom,
		]
		\addplot+ [blue, boxplot, line width=0.8pt, mark options={scale=0.8, opacity=0.3}] table [y=sharpnessvals]
			{pgfplots/data/uncorrelated_sharpness_lsq.dat};
		\addplot+ [purple, boxplot, line width=0.8pt, mark options={scale=0.8, opacity=0.3}] table [y=sharpnessvals]
			{pgfplots/data/uncorrelated_sharpness_lin.dat};
		\addplot+ [darkgreen, boxplot, line width=0.8pt, mark options={scale=0.8, opacity=0.3}] table [y=sharpnessvals]
			{pgfplots/data/uncorrelated_sharpness_mcmc.dat};

	\end{axis}

\end{tikzpicture}

%% file: pgfplots/calib_long_tail.tex
\begin{tikzpicture}

	\begin{axis}[
			name=plot1,
			footnotesize,
			enlargelimits=0.05,
			xlabel={\emph{Theoretical}\\[-0.75ex] \emph{coverage}},
			ylabel={\emph{Empirical}\\[-0.75ex] \emph{coverage}},
			x label style={at={(xticklabel cs:1)}, anchor=north east, inner sep=0pt, outer sep=0pt, align=right, scale=1.0},
			y label style={at={(yticklabel cs:1)}, rotate=-90, anchor=south west, align=left, scale=1.0},
			ytick distance=0.25,
			xtick align=center,
			ytick align=inside,
			axis background/.style={fill=bgcolor},
			line width=0pt,
			y tick style={line width=0pt},
			xtick pos=bottom,
			ymajorgrids=true,
			legend style={fill=bgcolor, inner sep=1pt, fill opacity=0.5, text opacity=1, nodes={scale=0.8}, draw=none},
			legend cell align={left},
			legend pos=north west,
		]

		\addplot [ blue, line width=1.0pt,]
		table[x=pvals, y=calibrationvals] {pgfplots/data/long_tail_calibration_lsq.dat}
		coordinate[pos=0.5] (foo)
		;
		\addlegendentry{LSQ};

		\addplot [ purple, line width=1.0pt,]
		table[x=pvals, y=calibrationvals] {pgfplots/data/long_tail_calibration_lin.dat}
		;
		\addlegendentry{Linear};

		\addplot [ darkgreen, line width=1.0pt,]
		table[x=pvals, y=calibrationvals] {pgfplots/data/long_tail_calibration_mcmc.dat}
		;
		\addlegendentry{MCMC};



		\addplot [
			gray,
			opacity=0.5,
			line width=0.8pt,
		]
		coordinates {(0,0) (1,1)};

	\end{axis}

	\begin{axis}[at=(plot1.north east), anchor=left of north west, xshift=0.20cm,
			name=plot2,
			y label style={at={(yticklabel cs:1)}, rotate=-90, anchor=south west, align=left, scale=1.0},
			x label style={at={(xticklabel cs:1)}, anchor=north east, inner sep=0pt, outer sep=0pt, align=right, scale=1.0, yshift=-0.33ex},
			ylabel={\emph{Sharpness}\\[-0.75ex] \emph{(\textdownarrow\ better)}},
			enlargelimits=0.05,
			footnotesize,
			boxplot/draw direction=y,
			axis background/.style={fill=bgcolor},
			xtick={1, 2, 3},
			xticklabels={\emph{LSQ}, \emph{Linear}, \emph{MCMC}},
			yticklabel={\pgfmathprintnumber{\tick} $\unit{\m}$},
			x tick label style={rotate=0},
			xtick align=inside,
			ytick align=inside,
			ytick style={line width=0pt},
			ymin=0,
			ymax=1000,
			ymajorgrids=true,
			line width=0pt,
			xtick align=center,
			xtick pos=bottom,
		]
		\addplot+ [ blue, boxplot, line width=0.8pt, mark options={scale=0.8, opacity=0.3}] table [y=sharpnessvals]
			{pgfplots/data/long_tail_sharpness_lsq.dat};
		\addplot+ [ purple, boxplot, line width=0.8pt, mark options={scale=0.8, opacity=0.3}] table [y=sharpnessvals]
			{pgfplots/data/long_tail_sharpness_lin.dat};
		\addplot+ [ darkgreen, boxplot, line width=0.8pt, mark options={scale=0.8, opacity=0.3}] table [y=sharpnessvals]
			{pgfplots/data/long_tail_sharpness_mcmc.dat};

	\end{axis}

\end{tikzpicture}

%% file: pgfplots/calib_filtered_x.tex
\begin{tikzpicture}[
		overlaytext/.style={fill=bgcolor, fill opacity=0.66, text opacity=1}
	]
	\begin{axis}[
			name=plot1,
			footnotesize,
			enlargelimits=0.05,
			xlabel={\emph{Theoretical}\\[-0.75ex] \emph{coverage}},
			ylabel={\emph{Empirical}\\[-0.75ex] \emph{coverage}},
			x label style={at={(xticklabel cs:1)}, anchor=north east, inner sep=0pt, outer sep=0pt, align=right, scale=1.0},
			y label style={at={(yticklabel cs:1)}, rotate=-90, anchor=south west, align=left, scale=1.0},
			xtick distance=0.25,
			ytick distance=0.25,
			xticklabel style={/pgf/number format/.cd,frac, frac TeX=\nicefrac},
			yticklabel style={/pgf/number format/.cd,frac, frac TeX=\nicefrac},
			xtick align=center,
			ytick align=inside,
			axis background/.style={fill=bgcolor},
			line width=0pt,
			y tick style={line width=0pt},
			xtick pos=bottom,
			ymajorgrids=true,
		]
		\addplot [ purple, line width=1.0pt,]
		table[x=pvals, y=calibrationvals] {pgfplots/data/calib_filtered_z.dat}
		node[left, pos=0.75, font=\scriptsize, inner sep=2pt, anchor=east] {altitude};
		;

		\addplot [ blue, line width=1.0pt,]
		table[x=pvals, y=calibrationvals] {pgfplots/data/calib_filtered_x.dat}
		coordinate[pos=0.5] (foo)
		node[left, pos=0.45, font=\scriptsize, align=center, yshift=2ex, name=bar] {along-\\track};
		\draw[blue, thin, shorten >={-4pt}] (foo) to (bar);

		\addplot [ darkgreen, line width=1.0pt,]
		table[x=pvals, y=calibrationvals] {pgfplots/data/calib_filtered_y.dat}
		node[below right, pos=0.60, font=\scriptsize, inner sep=0pt, anchor=north west, align=center] {cross-\\track};
		;


		\addplot [ violet, line width=1.0pt,]
		table[x=pvals, y=calibrationvals] {pgfplots/data/calib_filtered_joined.dat}
		node[below, pos=0.48, font=\scriptsize, inner sep=1pt, align=center] {joint};

		\addplot [
			gray,
			opacity=0.5,
			line width=0.8pt,
		]
		coordinates {(0,0) (1,1)};

	\end{axis}

	\begin{axis}[at=(plot1.north east), anchor=left of north west, xshift=0.20cm,
			name=plot2,
			y label style={at={(yticklabel cs:1)}, rotate=-90, anchor=south west, align=left, scale=1.0},
			ylabel={\emph{Sharpness}\\[-0.75ex] \emph{(\textdownarrow\ better)}},
			enlargelimits=0.05,
			enlarge y limits={upper=+0.14},
			footnotesize,
			boxplot/draw direction=y,
			axis background/.style={fill=bgcolor},
			xmin = -6000,
			xtick align=inside,
			ytick align=inside,
			xtick distance = 1000,
			xtick=\empty,
			xmax = -4000,
			ymin=20,
			ymax=400,
			ytick={100, 250, 400},
			ytick style={line width=0pt},
			height=3cm,
			ymajorgrids=true,
			line width=0pt,
			yticklabel={\pgfmathprintnumber{\tick} $\unit{\m}$},
		]
		\addplot[blue, densely dotted, line width=0.7pt] table [x=alongtrack_dist, y=sharpnessvals]  {pgfplots/data/marginal_sharpness_preds_1_1.dat} coordinate[pos=0.45] (label1) coordinate[pos=0.9] (c1);
		\addplot[blue, line width=0.7pt] table [x=alongtrack_dist, y=sharpnessvals]  {pgfplots/data/marginal_sharpness_filtered_1_1.dat} coordinate[pos=0.32] (label2) coordinate[pos=0.9] (c2);
		\draw[->, shorten <={1pt}, shorten >={1pt}, line width=0.75pt] (c1) to
		node[left=0.0em, font={\tiny\bfseries}, inner sep=0.15em, align=center, scale=1.0, xshift=-0.15em, yshift=-0.4ex, overlaytext]
			{$\bm{2 \times}$impro-\\ vement} (c2);
		\node[blue, align=left, font={\tiny\bfseries}, anchor=south west, scale=1.0, xshift=0.0em, yshift=0.0ex, inner sep=0.1em, overlaytext] at (label1) {unfiltered \\ alongtrack};
		\node[blue, align=left, font={\tiny\bfseries}, anchor=north east, scale=1.0, xshift=0.0em, yshift=0.8ex, overlaytext] (filtered) at (label2) {filtered};
		\node[below=-0.0 of filtered.south west, blue, align=left, font={\tiny\bfseries}, anchor=north west, scale=1.0, overlaytext, yshift=0.0ex] (filtered) {alongtrack};




	\end{axis}

	\begin{axis}[at=(plot2.south), anchor=north, yshift=-0.1cm,
			name=plot2,
			enlargelimits=0.05,
			enlarge y limits={lower=+0.15},
			footnotesize,
			boxplot/draw direction=y,
			axis background/.style={fill=bgcolor},
			xmin = -6000,
			xtick distance = 1000,
			ytick distance = 15,
			scaled x ticks=base 10:-3,
			xtick scale label code/.code={},
			xticklabel={\pgfmathprintnumber{\tick} km},
			yticklabel={\pgfmathprintnumber{\tick} $\unit{\m}$},
			y tick style={line width=0pt},
			ytick align=inside,
			xtick pos=bottom,
			xtick align=center,
			xmax = -4000,
			ymin=0,
			height=2.8cm,
			xlabel={\emph{Alongtrack}\\[-0.75ex] \emph{position}},
			x label style={at={(xticklabel cs:1)}, anchor=north east, inner sep=0pt, outer sep=0pt, align=right, scale=1.0, yshift=-0.33ex},
			ymajorgrids=true,
			line width=0pt,
		]

		\addplot[darkgreen, densely dotted, line width=0.7pt] table [x=alongtrack_dist, y=sharpnessvals]  {pgfplots/data/marginal_sharpness_preds_2_2.dat}
		coordinate[pos=0.0] (label3) coordinate[pos=0.85] (c3);
		\addplot[darkgreen, line width=0.7pt] table [x=alongtrack_dist, y=sharpnessvals]  {pgfplots/data/marginal_sharpness_filtered_2_2.dat}
		coordinate[pos=0.40] (label4) coordinate[pos=0.85] (c4);

		\addplot[purple, densely dotted, line width=0.7pt] table [x=alongtrack_dist, y=sharpnessvals]  {pgfplots/data/marginal_sharpness_preds_3_3.dat}
		coordinate[pos=0.44] (label5) coordinate[pos=0.8] (c5);
		\addplot[purple, line width=0.7pt] table [x=alongtrack_dist, y=sharpnessvals]  {pgfplots/data/marginal_sharpness_filtered_3_3.dat}
		coordinate[pos=0.28] (label6) coordinate[pos=0.8] (c6);
		\node[purple, align=left, font={\tiny\bfseries}, anchor=south west, scale=1.0, xshift=0.0em, yshift=-0.0ex, inner sep=1.0pt,
			overlaytext]
		at (label6) {altitude};
		\draw[->, shorten <={1pt}, shorten >={1pt}, line width=0.75pt]
		(c5) to coordinate[right] (2ximpr)
		(c6);

		\node[green!50!black, align=left, font={\tiny\bfseries}, anchor=north east, scale=1.0, xshift=0.0em, yshift=0.0ex, inner sep=0.1pt, overlaytext] at (label4) {crosstrack};
		\draw[->, shorten <={0.5pt}, shorten >={0.5pt}, line width=0.75pt]
		(c3) to
		coordinate[right] (2ximpr2)
		(c4);

		\node[font={\tiny\bfseries}, scale=1.0, align=center, inner sep=0.1em, overlaytext] (2xnode) at (axis cs: -4250,26) {$\bm{2 \times}$impro-\\ vement};
		\draw[shorten <={-0pt}, shorten >={2pt}, line width=0.5pt, darkgray] (2xnode) to (2ximpr);
		\draw[shorten <={-0pt}, shorten >={2pt}, line width=0.5pt, darkgray] (2xnode) to[bend left=10] (2ximpr2);


	\end{axis}

\end{tikzpicture}

%% file: sections/conclusion.tex
\section{Conclusion}

We presented how to incorporate knowledge of measurement
uncertainties into a parameter estimation pipeline to yield a
distribution instead of a point estimate, and applied this to
aircraft pose estimation using image features corresponding to known
runway features, although other geometric properties such as sideline
angles can be incorporated just the same. To this end, we have presented
three estimators with trade-offs in computational speed, mathematical
assumptions, and ease of implementation.

We have also highlighted that evaluating and comparing probabilistic
estimators is no simple task and requires analyzing calibration and
sharpness properties, which are not obvious for multivariate
predictions. To this end, we have put forth a new definition of
calibration and sharpness specifically for multivariate normal
distributions.

We have used this new definition to compare the three estimators given
different noise models and have found that the Linear Approximation
Estimator performs on par with the other two estimators when the noise
is Gaussian with known covariance. However, for non-Gaussian noise, the
Linear Approximation Estimator performs poorly and is outperformed in
particular by the MCMC-based approach, which still manages to produce
well-calibrated and sharp results. The MCMC-based approach is faster than the LSQ-based
approach while also being easy to implement.

Finally, we have shown that the estimator outputs can be used as inputs
to a Kalman filter using a case study with real measurements, improving
sharpness by about \(2 \times\) while maintaining approximate marginal
calibration for all three position variables. Somewhat surprisingly,
filtering seems to introduce errors in the covariate terms of the
predictions, which leads to poor calibration of the joint distributions.

This work aims to aid adoption of machine learning-based
sensors into safety-critical applications by enabling rigorous
integration and careful evaluation of uncertainty estimates coming from
the sensor. Further, we hope to give guidance to practitioners trying to
choose a suitable estimator for probabilistic parameter estimation and
provide reference implementations in the Julia language.

%% file: sections/acknowledgements.tex
\section*{Acknowledgements}
The authors would like to thank A$^3$ by Airbus for their generous funding and support of this research.
We are in particular grateful to Kinh Tieu, Jit Chowdhury, and Mansur Arief for their valuable insights and discussions throughout the course of this work.
Additionally, we wish to acknowledge the developers of the SciML, Turing.jl, and Julia communities for providing the ecosystem upon which we have built this work.


%% file: sections/appendix.tex
\appendix



\subsection{Experimental Setup: Runway and Aircraft}\label{subsec:runway-details}
We consider a runway with \(\qty{3500}{\metre}\) length and \(\qty{60}{\metre}\) width. For each
prediction, we sample a random camera position from a cone
approaching the runway, with
\begin{equation}\label{eq:random-pose}
	\begin{aligned}
		\beta^{\text{alongtrack }}                                   & \sim U\lbrack - \qty{6000}{\metre}, - \qty{4000}{\metre}\rbrack                                                                     \\
		{\beta^{\text{crosstrack }}}                                 & \sim U\left\lbrack - \tan(\qty{20}{\degree}),\tan(\qty{20}{\degree}) \right\rbrack \cdot {\left| \beta^{\text{alongtrack}} \right|} \\
		{\beta^{\text{altitude }}}                                   & \sim U\left\lbrack \tan(\qty{1}{\degree}),\tan(\qty{2}{\degree}) \right\rbrack \cdot {\left| \beta^{\text{alongtrack}} \right|}     \\
		\beta^{\text{roll}},\beta^{\text{pitch}},\beta^{\text{yaw }} & \sim U\lbrack - \qty{10}{\degree},\qty{10}{\degree}\rbrack
	\end{aligned}
\end{equation}
where \(U\lbrack a,b\rbrack\) denotes a uniform
distribution over the interval \(\lbrack a,b\rbrack\). After sampling a
true pose \(\bbeta^{\ast}\), we construct a prior distribution
\begin{gather}
	p_{{\bbetah}} = \bcalN_{\left( \beta^{\ast},\Sigma_{\bbeta} \right)} \\
	\Sigma_{\bbeta} = \text{diag}\left( (\qty{1000}{\metre})^{2},(\qty{200}{\metre})^{2},(\qty{200}{\metre})^{2} \right)
	\label{eq:beta-prior}
\end{gather}
which is used as the prior for the MCMC Estimator and from which the initial guesses are
sampled for the LSQ and Linear Approximation Estimators.

\subsection{Solver Details: LSQ Estimator and Linear Approximation
	Estimator}\label{subsec:lsq-details}

We solve the nonlinear least squares problem \cref{eq:lsq4} using the trust region implementation provided by
\texttt{NonlinearSolve.jl} framework \cite{palNonlinearSolveJlHighperformance2024} with default parameters.
The derivatives, both for the solver and for computing the Jacobian in \cref{eq:linearmodel1}, are computed using the \texttt{ForwardDiff.jl} framework \cite{revelsForwardModeAutomaticDifferentiation2016}.
For the each experiment for the LSQ Estimator, we compute \(400\) samples of \(\bbetah\) before fitting the normal distribution.
For both estimators, we sample an initial guess for
\(\bbeta^{(k = 1)}\) from the prior distribution
\(p_{\bbetah}\) given in \cref{eq:beta-prior}.
For the LSQ Estimator, starting from \(k = 2\), we use the solution of
the previous sample as the initial guess.

\subsection{Solver Details: MCMC Estimator}\label{subsec:mcmc-details}

We implement the probabilistic program given in \cref{eq:mcmc2} using the
\texttt{Turing.jl} framework \cite{ge2018turing}. We use the No-U-Turn-Sampler
algorithm~\cite{hoffmanNoUTurnSamplerAdaptively2014} with acceptance rate
\(0.65\) but have also found good results with other Hamiltonian Monte Carlo algorithms, but not with simpler algorithms like Metropolis-Hastings.
We use forward-mode automatic differentiation through the
\texttt{ForwardDiff.jl} framework~\cite{revelsForwardModeAutomaticDifferentiation2016} to compute the required gradients.

For the prior \(p_{\bbetah}\), we use \cref{eq:beta-prior}.
We find that for poor priors (too narrow or at a bad location) the sharpness deteriorates, although the predictions stay well calibrated.

When sampling, we find that we need at least 250 steps of ``burn in'' samples which
are discarded.
Similar to the LSQ Estimator, we
generate 400 samples to fit the multivariate normal distribution, thereby generating 650 samples in total.